\documentclass{llncs}

\usepackage{graphicx}
\usepackage{amsmath,amssymb,stmaryrd,textcomp}
\usepackage{proof}

\title{I/O Logic in HOL --- First Steps}
\author{Christoph Benzm\"uller \and Xavier Parent}
\institute{Computer Science and Communications, University of Luxembourg, Luxembourg} 

\begin{document}
\maketitle

\begin{abstract} A semantical embedding of input/output logic in classical higher-order logic is presented. This embedding enables the mechanisation and automation of reasoning tasks in input/output logic with off-the-shelf higher-order theorem provers and proof assistants. The key idea for the solution presented here results from the analysis of an inaccurate previous embedding attempt, which we will discuss as well.
\end{abstract}

\section{Input/Output Logic}
Input/output (I/O) logic, initially devised by Makinson 
\cite{DBLP:journals/jphil/MakinsonT00} and  further developed by Leon van der Torre and colleagues, is an deontic logic framework that has gained increased recognition in the AI
community. This is evidenced by the fact that the framework has its
own chapter in the handbook of deontic logic and
normative systems \cite{gabbay13:_handb_deont_logic_normat_system}.

I/O logic operators, such as the (\emph{simple-minded}) output operation $out_1$, accept a set $G$ of conditional norms as argument. The conditional norms in $G$ are given as pairs $(a,x)$. The body $a$ of such a pair, called the input, is representing some condition or situation. The head $x$, called output, is representing what is desirable or obligatory in the given situation. 
$a$ and $x$ are thereby propositional formulas. A key point is that the pair $(a,x)$ is not given a truth-functional semantics in I/O logic. 

Different kinds of I/O operations have been presented in the literature. In this short note we focus on the I/O logic operator $out_1$. Future work will extend the presented work to other  I/O operations. When $out_1$ is applied to a set $G$ of conditional norms and a set $A$ of propositions describing of an input situation, it tells us what is desirable or obligatory  in this situation according to $G$.
 
The \emph{semantics} of I/O logic operator $out_1$ is defined as follows ($P$ is the set of propositional formulas and $\models$ is the associated propositional logic consequence relation): 
$out_1(G,A) := Cn(G(Cn(A)))$, with $Cn(X) := \{s \in P  \mid X \models s\}$  and $G(X) := \{s \in P \mid \text{ there exists } a \in X \text{ such that } (a,s) \in G \}$. 
For technical reasons we below restrict $out_1(G,a)$ to operate only on a single propositional input formula $a$. However, one may consider $a$ as the conjunction of all formulas $a_i\in A$. 

A \emph{proof theory} for $out_1$ is given by the following proof rules:
\[
\mbox{\infer[\textit{SO}]{(a,c)}{(a,b) & b \models c}} \qquad 
\mbox{\infer[\textit{WI}]{(a,c)}{(b,c) & a \models b}}  \qquad 
\mbox{\infer[\wedge]{(a,b \wedge c)}{(a,b) & (a,c)}} \qquad
\mbox{\infer[\top]{(\top,\top)}{}}
\]

This proof system works as follows. In order to check whether $x$ in $out_1(G,a)$, the proof system is first expanded by adding a rule \infer[]{(b,c)}{} for each pair $(b,c)\in G$. Then it is checked whether the pair $(a,x)$ can be derived 
from finitely many pairs in G using the rules above.

\section{Embedding of I/O Logic in HOL -- Initial Attempt} \label{sec2}
\begin{figure}[t] \centering
\includegraphics[width=.9\textwidth]{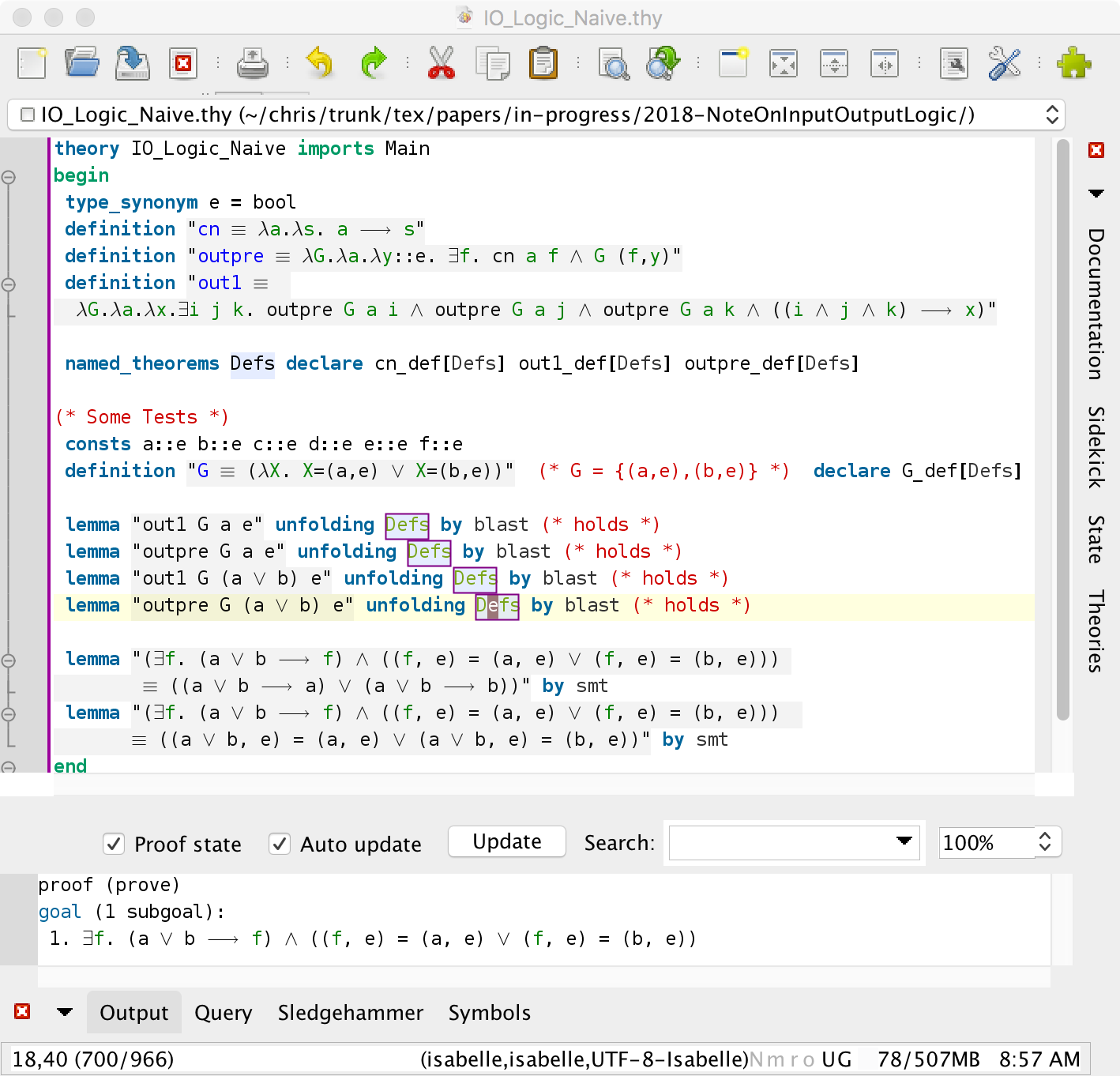}
\caption{Unsound, naive embedding of I/O logic in HOL \label{fig1}}
\end{figure}

An initial, naive attempt to embed I/O logic in Isabelle/HOL \cite{Isabelle} did fail. However, a careful failure analysis was key for devising the proper embedding as presented in Sec. \ref{sec3} below. We briefly recap this initial embedding attempt, which is also displayed in Fig.~\ref{fig1}, and explain the issue. 

This section and the remainder of this paper requires some background knowledge about HOL. However, due to lack of space we do not present a respective introduction here and instead refer to the literature \cite{J6,B5,AndrewsSEP}.


\sloppy In our naive embedding attempt the statement $a \models s$ was simply mapped to implication:
$\models_{o \shortrightarrow o\shortrightarrow o}\ :=\ \lambda  a_o \lambda s_o (a \supset s)$. This was inspired by the deduction theorem for classical propositional logic: $a \models s$ iff $\models a \supset s$. Subsequently, an operation $outpre$ was defined as 
$outpre_{((o \times o) \shortrightarrow o) \shortrightarrow o \shortrightarrow (o\shortrightarrow o)} := \lambda G_{(o \times o) \shortrightarrow o} \lambda a_o \lambda y_o \exists f_o ((Cn\, a) f \wedge G (f,y))$, the idea being that $outpre(N,a)$ should denote the set $\{y \in P \mid \text{ exists } f \in Cn(a) \text{ such that } (f,y) \in N\}$, that is $outpre(G,a) = G(Cn(a))$.
In a final step, a pragmatically motivated approximation\footnote{In this experimental phase this approximation was sufficient and a proper definition was postponed for later. In fact, this approximation is not influencing the core problem as discussed below.} of ${out_1}$ was defined as ${out_1}_{((o \times o) \shortrightarrow o) \shortrightarrow o \shortrightarrow (o\shortrightarrow o)}:= \lambda G_{(o \times o) \shortrightarrow o} \lambda a_o \lambda x_o 
\exists h_o \exists i_o \exists j_o ((outpre\ N\ a\ h) \wedge (outpre\ N\ a\ i) \wedge (outpre\ N\ a\ j) \wedge ((h \wedge i \wedge j) \supset x))$. That is, $out'_1(N,a)$ denotes the set $\{x \in P \mid \{a,j,k\} \subseteq outpre(N,a) \wedge 
((h \wedge i \wedge j) \supset x)$. This is an approximation of $out_1$ in the sense that not all consequences of $outpre(N,a)$ are modelled, but only those that follow from maximally three formulas in $outpre(N,a)$. Moreover, to keep the discussion as simple as possible we assume here that $outpre(N,a)$ is non-empty.

Two simple, running examples are used in the remainder of this section to illustrate the fundamental problem with this naive embedding attempt. Let the set of conditional norms $G$ be given as $\{(a,e),(b,e)\}$, for propositional symbols $a$, $b$, and $e$.  Example E1 asks whether \emph{$e$ is in $out_1(G,a)$}, and example E2 asks whether \emph{$e$ is in $out_1(G,a\vee b)$}.  The former is expected to hold, but the latter not (for simple minded output). However, when utilising the above embedding both E1 and E2 unfold into valid HOL formulas. The reason for this unsound behaviour can be well explained already on the basis of $outpre$ alone, which is preferable since it leads to smaller unfolded HOL formulas we need to discuss. Thus, we modify the examples into E1': \emph{$e$ is in $outpre(G,a)$} and E2: \emph{$e$ is in $outpre(G,a\vee b)$}.

Unfolding the embedding for E1' results in the HOL formula 
$\exists f. (a \supset f) \wedge ((f,e) = (a,e) \vee (f,e) = (b,e))$, which is valid as intended: simply choose $f$ as $a$.
Unfolding E2' analogously results in $\exists f. ((a \vee b) \supset f) \wedge ((f,e) = (a,e) \vee (f,e) = (b,e))$. However, contrary to our intention, this formula is also valid in HOL. To see this, choose $f$ as $a\vee b$: $((a \vee b) \supset (a \vee b)) \wedge (((a \vee b),e) = (a,e) \vee ((a \vee b),e) = (b,e))$. By simplification, Boolean extensionality and congruence this formula is equivalent to\footnote{This formula is also equivalent to $(a\vee b,e) = (a,e) \vee (a\vee b,e) = (b,e)$. For proving this alternative, simplified formula we proceed as follows. We know that $(a\vee b) \vee \neg (a\vee b)$ holds by the law of excluded middle. We proceed by case distinction. If $a\vee b$ is true we have $a$ is true or $b$ is true. In the former case we get $(a\vee b,e) = (a,e)$ and in the latter case we have $(a\vee b,e) = (b,e)$, and we are done. If $\neg(a\vee b)$ is false, we have $a$ is false and $b$ is false, and thus both $(a\vee b,e) = (a,e)$ and $(a\vee b,e) = (b,e)$ are true by Boolean extensionality and congruence.}
$((a\vee b) \supset a) \vee ((a\vee b) \supset b)$, which is valid in classical logic: We know that $(a\vee b) \vee \neg (a\vee b)$ holds by the law of excluded middle. Hence, we proceed by case distinction. If $a\vee b$ is true, we have $a$ is true or $b$ is true. In both cases the statement follows trivially.  The statement is also trivially true in case $a\vee b$ is false. 

We have thus reduced E2': \emph{$e$ is in $outpre(G,a\vee b)$} to $\models ((a\vee b) \supset a) \vee ((a\vee b) \supset b)$. Intuitively, however, we should have reduced E2' to (($\models (a\vee b) \supset a$) or  ($\models (a\vee b) \supset a$)), which is not equivalent, since the former does not imply the latter.

The problem obviously is this: $a \models s$, respectively $\models a \supset s$, cannot be simply encoded as $a \supset b$, at least not when this formula is subsequently nested in other  formulas, as done here in the definition of $outpre$. Such an encoding and nested usage, in combination with the law of excluded middle, causes the observed unsound behaviour. 
A more appropriate modeling of $\models a \supset s$ is thus required.

\section{Proper Embedding of I/O Logic in HOL} \label{sec3}
\begin{figure}[t]
\includegraphics[width=.9\textwidth,height=11.6cm]{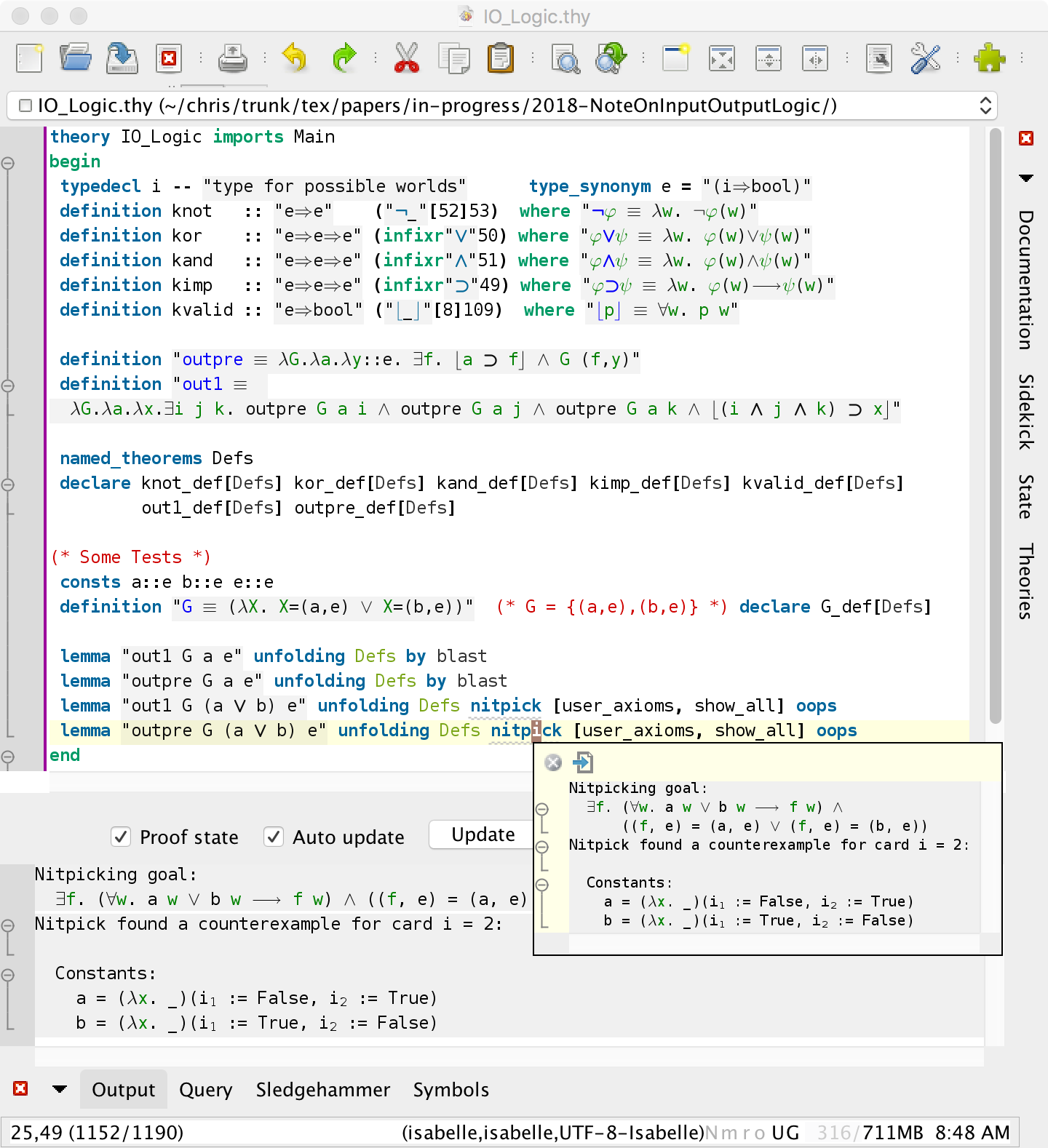} \centering
\caption{Proper embedding of I/O logic in HOL. (Note: We still assume here that $outpre(G,a)$ is non-empty. To include the cases where $outpre(G,a)$ is empty it suffices to add a disjunct 
``$\, \vee \, \lfloor x \rfloor$''  to the existential statement in the definition of $out1$.) \label{fig2}}
\end{figure}

To obtain a proper embedding of I/O Logic in HOL
we devise a suitable encoding of $\models \varphi$, cf. the Isabelle/HOL encoding of our solution in Fig.~\ref{fig2}.
By suitable we mean that the new encoding of $\models \varphi$ can be nested in larger formula contexts without causing the effect as discussed in Sec.~\ref{sec2}. This can be achieved by lifting the propositional formulas of I/O logic to predicates on possible worlds (or states). We thus reuse a technique from previous work in which the objective was to properly embed (quantified) modal logics in HOL \cite{J23}. However, the reason for applying this technique is different here. Our challenge is not to properly encode e.g. the modal box operator, but to introduce a suitable encoding of $\models a \supset s$, so that occurrences of this term can be soundly nested in other formulas, while blocking certain undesirable effects of the law of excluded middle.

Propositional formulas $\varphi$ of I/O logic therefore mapped to associated HOL predicates of type $i\rightarrow o$, where type $i$ denotes a (non-empty) set of possible worlds (or states). The logical connectives $\boldsymbol{\neg}$, $\boldsymbol{\vee}$, $\boldsymbol{\wedge}$ and $\boldsymbol{\supset}$ of I/O logic defined in this mapping as follows: $\boldsymbol{\neg} \varphi := \lambda w \neg (\varphi w)$, $\varphi \boldsymbol{\vee} \psi := \lambda w (\varphi w \vee \psi w)$, $\varphi \boldsymbol{\wedge} \psi := \lambda w (\varphi w \wedge \psi w)$, and $\varphi \boldsymbol{\supset} \psi := \lambda w (\varphi w \supset \psi w)$. The lifted I/O formulas in HOL are \emph{``grounded''} again to Boolean type by the following definition of validity: ${valid}\  \varphi := \forall w (\varphi w)$. The claim now is that $valid\ \varphi$, denoted in the remainder (and in related papers) also as $\lfloor \varphi \rfloor$, suitably encodes $\models \varphi$ ins such a way that this term can safely be nested in larger formula context without causing the effects as observed in Sec.~\ref{sec2}. The previous definitions of $outpre$ is thus changed as follows: $outpre := \lambda G \lambda a \lambda y \exists f (\lfloor a \boldsymbol{\supset} f\rfloor \wedge G (f,y))$. The approximative encoding of $out_1$, which refers to $outpre$, remains unchanged.

We once again analyse the running examples, but now for the modified semantical embedding. Example E1': \emph{$e$ is in $outpre(G,a)$} now unfolds into the HOL formula $\exists f ((\forall w (a w \supset f w)) \wedge ((f, e) = (a, e) \vee (f, e) = (b, e)))$. This formula is valid as intended: simply instantiate $f$ with $\lambda x (a x)$. Similarly, example E1: \emph{$e$ is in $out_1(G,a)$} unfolds into a valid HOL formula, and we leave the details of this example to the reader.

The more interesting example E2': \emph{$e$ is in $outpre(G,a\vee b)$} now unfolds into $\exists f ((\forall w. (a w \vee b w) \supset f w) \wedge  ((f, e) = (a, e) \vee (f, e) = (b, e))$. We apply an analogous idea as before and instantiate $f$ with $\lambda x (a x \vee b x)$, which results, after normalisation and simplification, in the HOL formula 
$(\lambda x (a x \vee b x)), e) = (a, e) \vee ((\lambda x (a x \vee b x)), e) = (b, e)$. Contrary to the situation in Sec.~\ref{sec2}, the law of excluded middle cannot be exploited anymore to prove this formula.
This formula in fact has a countermodel:  Consider two possible worlds $i_1$ and $i_2$, and choose $a$ as the set $\{i_2\}$ (i.e., $a$ is true only in world $i_2$) and $b$ as the set $\{i_1\}$; then $\lambda x (a x \vee b x)$ denotes the set $\{i_1, i_2\}$, and our formula obviously evaluates to false. 
We obtain respective countermodels for both  E2' and E2, and these countermodels are quickly found automatically by the model finder Nitpick \cite{BN10} as illustrated in Fig.~\ref{fig2}.

\section{Conclusion}
\begin{figure}[t]
\includegraphics[width=.9\textwidth,height=12.6cm]{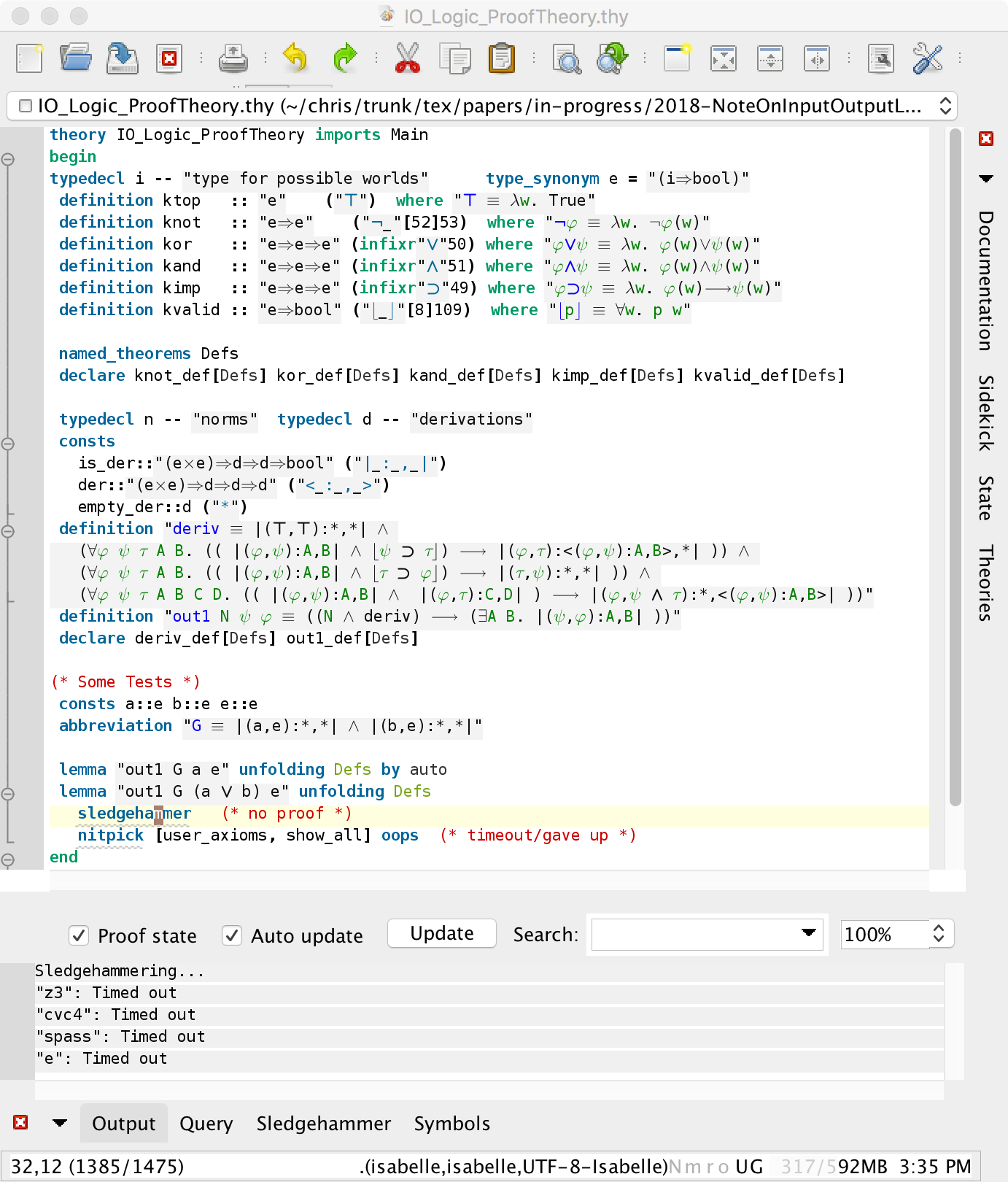} \centering
\caption{Embedding of the proof theory of I/O logic in HOL \label{fig3}}
\end{figure}

Just in time for the 50st birthday of Leon van der Torre we have devised a proper semantical embedding of the simple minded I/O logic operation $out_1$ into HOL. Further work includes the formal validation of the faithfulness (soundness and completeness) of the embedding, and extensions to further I/O operations.

Ongoing work is utilising the very same technique as employed in this paper for the embedding of the proof theory of I/O logic in HOL. The current status of these activities is depicted in Fig.~\ref{fig3}. An advantage is that an approximation of $out_1$ can easily be avoided. A disadvantage, however, is that proof automation and countermodel finding seems to become less effective, this is what current experiments indicate. Due the cut-introducing nature of the I/O proof rules \textit{SO} and \textit{WI}, when applied in backward direction, this is not so surprising though.

Another promising direction for future work is to devise an alternative semantical embedding of I/O logic in HOL based on Parent's recent interpretation of I/O Logic in intuitionistic logic \cite{Parent2014}. This work could be based on the already existing semantical embedding of intuitionistic logic in HOL~\cite{J21}.

\bibliographystyle{abbrv}
\bibliography{CJL}

\begin{thebibliography}{10}

\bibitem{AndrewsSEP}
P.~Andrews.
\newblock Church's type theory.
\newblock {\em In: E.N. Zalta (ed.) The Stanford Encyclopedia of Philosophy},
  Spring, 2014.

\bibitem{J6}
C.~Benzm{\"u}ller, C.~Brown, and M.~Kohlhase.
\newblock Higher-order semantics and extensionality.
\newblock {\em Journal of Symbolic Logic}, 69(4):1027--1088, 2004.

\bibitem{B5}
C.~Benzm{\"u}ller and D.~Miller.
\newblock Automation of higher-order logic.
\newblock In D.~M. Gabbay, J.~H. Siekmann, and J.~Woods, editors, {\em Handbook
  of the History of Logic, Volume 9 --- Computational Logic}, pages 215--254.
  North Holland, Elsevier, 2014.

\bibitem{J21}
C.~Benzm{\"u}ller and L.~Paulson.
\newblock Multimodal and intuitionistic logics in simple type theory.
\newblock {\em The Logic Journal of the IGPL}, 18(6):881--892, 2010.

\bibitem{J23}
C.~Benzm{\"u}ller and L.~Paulson.
\newblock Quantified multimodal logics in simple type theory.
\newblock {\em Logica Universalis (Special Issue on Multimodal Logics)},
  7(1):7--20, 2013.

\bibitem{BN10}
J.~C. Blanchette and T.~Nipkow.
\newblock Nitpick: A counterexample generator for higher-order logic based on a
  relational model finder.
\newblock In M.~Kaufmann and L.~C. Paulson, editors, {\em ITP 2010}, volume
  6172 of {\em LNCS}, pages 131--146. Springer, 2010.

\bibitem{gabbay13:_handb_deont_logic_normat_system}
D.~Gabbay, J.~Horty, X.~Parent, R.~van~der Meyden, and L.~van~der Torre,
  editors.
\newblock {\em Handbook of Deontic Logic and Normative Systems}.
\newblock College Publications, 2013.

\bibitem{DBLP:journals/jphil/MakinsonT00}
D.~Makinson and L.~W.~N. van~der Torre.
\newblock Input/output logics.
\newblock {\em J. Philosophical Logic}, 29(4):383--408, 2000.

\bibitem{Isabelle}
T.~Nipkow, L.~C. Paulson, and M.~Wenzel.
\newblock {\em Isabelle/HOL --- A Proof Assistant for Higher-Order Logic},
  volume 2283 of {\em LNCS}.
\newblock Springer, 2002.

\bibitem{Parent2014}
X.~Parent, D.~Gabbay, and L.~van~der Torre.
\newblock An intuitionistic basis for input/output logic.
\newblock In S.~O. Hansson, editor, {\em David Makinson on Classical Methods
  for Non-Classical Problems}, volume~3 of {\em Oustanding Contributions to
  Logic}, pages 263--286. Springer, 2014.

\end{thebibliography}

\end{document}